# $d$-SEPARATION: FROM THEOREMS TO ALGORITHMS

Dan Geiger, Thomas Verma & Judea Pearl
Cognitive Systems Laboratory, Computer Science Department
University of California Los Angeles, CA 90024
geiger@cs.ucla.edu

## ABSTRACT

An efficient algorithm is developed that identifies **all** independencies implied by the topology of a Bayesian network. Its correctness and maximality stems from the soundness and completeness of $d$-separation with respect to probability theory. The algorithm runs in time $O(|E|)$ where $E$ is the number of edges in the network.

## 1. INTRODUCTION

Bayesian networks encode properties of a probability distribution using directed acyclic graphs (dags). Their usage is spread among many disciplines such as: Artificial Intelligence [Pearl 1988], Decision Analysis [Howard and Matheson 1981; Shachter 1988], Economics [Wold 1964], Genetics [Wright 1934], Philosophy [Glymour et al. 1987] and Statistics [Lauritzen and Spiegelhalter 1988; Smith 1987]. A Bayesian network is a pair $(D, P)$ where $D$ is a dag and $P$ is a probability distribution called the *underlying* distribution. Each node $i$ in $D$ corresponds to a variable $X_i$ in $P$, a set of nodes $I$ correspond to to a set of variables $\mathbf{X}_I$ and $x_i, x_I$ denotes values drawn from the domain of $X_i$ and from the (cross product) domain of $\mathbf{X}_I$, respectively.[1] Each node in the network is regarded as a storage cell for the distribution $P(x_i \mid x_{\pi(i)})$ where $\mathbf{X}_{\pi(i)}$ is a set of variables that correspond to the parent nodes $\pi(i)$ of $i$. The underlying distribution represented by a Bayesian network is composed via

$$P(x_1, \cdots, x_n) = \prod_{i=1}^{n} P(x_i \mid x_{\pi(i)}), \qquad (1)$$

(when $i$ has no parents, then $\mathbf{X}_{\pi(i)} = \varnothing$). The role of a Bayesian network is to record a state of knowledge $P$, to provide means for updating the knowledge as new information is accumulated and to facilitate query answering mechanisms for knowledge retrieval [Lauritzen and Spiegelhalter 1988; Pearl 1988]. A standard query for a Bayesian network is to find the posterior distribution of a hypothesis variable $X_l$, given an evidence set $\mathbf{X}_J = x_J$ i.e., to compute $P(x_l \mid x_J)$ for each value of $X_l$ and for a given combination of values of $\mathbf{X}_J$. The answer to such queries can, in principle, be computed directly from equation (1) because this equation defines a full probability distribution. However, treating the underlying distribution as a large table instead of a composition of several small ones, might be very inefficient both in time and space requirements, unless we exploit independence relationships encoded in the network. To better understand the improvements and limitations that more efficient algorithms can achieve, the following two problems must be examined: Given a variable $X_k$, a Bayesian network $D$ and the task of computing $P(x_l \mid x_J)$; determine, without resorting to numeric calculations: 1) whether the answer to the query is sensitive to the value of $X_k$, and 2) whether the answer to the query is sensitive to the *parameters* $p_k = P(x_k \mid x_{\pi(k)})$ stored at node $k$. The answer to both questions can be given in terms of conditional independence. The value of $X_k$ does not affect this query if $P(x_i \mid x_J) = P(x_i \mid x_J, x_k)$ for all values of $x_i, x_k$ and $x_J$, or equivalently, if $X_i$ and $X_k$ are *conditionally independent* given $\mathbf{X}_J$, denoted by $I(X_i, \mathbf{X}_J, X_k)_P$. Similarly, whether the parameters $p_k$ stored at node $k$ would not affect the query $P(x_l \mid x_J)$ also reduces to a simple test of conditional independence, $I(X_i, X_J, \pi_k)$, where $\pi_k$ is a (dummy) parent node of $X_k$ representing the possible values of $p_k$.

---

*This work was partially supported by the National Science Foundation Grant #IRI-8610155 and Naval Research Laboratory Grant #N00014-89-J-2057.

(1) Note that bolds letters denote sets of variables.

The main contribution of this paper is the development of an efficient algorithm that detects these independencies directly from the topology of the network, by merely examining the paths along which $i$, $k$ and $J$ are connected. The proposed algorithm is based on a graphical criteria, called d-separation, that associates the topology of the network to independencies encoded in the underlying distribution. The main property of d-separation is that it detects only genuine independencies of the underlying distribution [Verma and Pearl 1988] be sharpened to reveal additional independencies [Geiger and Pearl 1988]. and that it can not A second contribution of the paper is providing a unified approach to the solution of two distinct problems: sensitivity to parameter values and sensitivity to variable instantiations.

## 2. SOUNDNESS AND COMPLETENESS OF $d$-SEPARATION

In this section we review the definition of $d$-separation; a graphical criteria that identifies conditional independencies in Bayesian networks. This criteria is both *sound* and *complete (maximal)* i.e., it identifies only independencies that hold in every distribution having the form (1), and all such independencies. A preliminary definition is needed.

**Definition:** A *trail* in a dag is a sequence of links that form a path in the underlying undirected graph. A node $\beta$ is called a head-to-head node with respect to a trail $t$ if there are two consecutive links $\alpha \rightarrow \beta$ and $\beta \leftarrow \gamma$ on $t$. (note that nodes that start and end a trail $t$ are not head-to-head nodes wrt to $t$).

**Definition [Pearl 1988]:** If $J$, $K$, and $L$ are three disjoint subsets of nodes in a dag $D$, then $L$ is said to $d$-separate $J$ from $K$, denoted $I(J,L,K)_D$, iff there is no trail $t$ between a node in $J$ and a node in $K$ along which (1) every head-to-head node (wrt $t$) either is or has a descendent in $L$ and (2) every node that delivers an arrow along $t$ is outside $L$. A trail satisfying the two conditions above is said to be *active*, otherwise it is said to be *blocked* (by $L$).

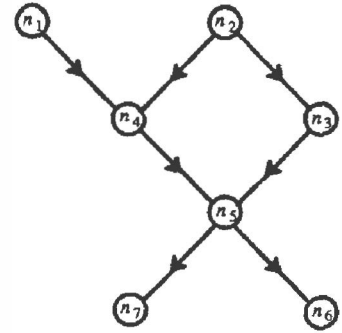

Figure 1

In Figure 1, for example, $J = \{n_4\}$ and $K = \{n_3\}$ are $d$-separated by $L = \{n_2\}$; the path $n_4 \leftarrow n_2 \rightarrow n_3$ is blocked by $n_2 \in L$ while the path $n_4 \rightarrow n_5 \leftarrow n_3$ is blocked because $n_5$ and all its descendents are outside $L$. Thus $I(n_4, n_2, n_3)_D$ holds in $D$. However, $J$ and $K$ are not $d$-separated by $L' = \{n_2, n_6\}$ because the path $n_4 \rightarrow n_5 \leftarrow n_3$ is rendered active: learning the value of the consequence $n_6$, renders its causes $n_3$ and $n_4$ dependent, like opening a pathway along the converging arrows at $n_5$. Consequently, $I(n_4, \{n_2,n_6\}, n_3)_D$ does not hold in $D$.

Note that in principle, to check whether $L$ $d$-separates $J$ and $K$, the definition requires an examination of all trails connecting a node in $J$ and a node in $K$, including trails that form a cycle in the underlying undirected graph. For example, in Figure 1, to check whether $J = \{n_1\}$ and $K = \{n_7\}$ are $d$-separated by $L = \{n_6\}$ would require checking trails such as $n_1, n_4, n_5, n_3, n_2, n_4, n_5, n_7$, and many others. The next lemma shows that a trail that forms a (undirected) loop need not be examined because whenever there is an active trail with a loop there is an active *simple* trail as well, i.e. a trail that forms no cycles in the underling undirected graph. In the previous example, the trail $n_1, n_4, n_5$ and $n_7$ is the simple active trail (by $\{n_6\}$), guaranteed by Lemma 1. The proof of lemma 1, which requires only the definition of d-separation, can be found in [Geiger at al. 1988].

**Lemma 1:** Let $L$ be a set of nodes in a dag $D$, and let $\alpha, \beta \notin L$ be two additional nodes of $D$. Then $\alpha$ and $\beta$ are connected via an active trail (by $L$) iff $\alpha$ and $\beta$ are connected via a simple active trail (by $L$).

**Definition:** If $\mathbf{X}_J$, $\mathbf{X}_K$, and $\mathbf{X}_L$ are three disjoint subsets of variables of a distribution $P$, then $\mathbf{X}_J$ and $\mathbf{X}_K$ are said to be conditionally independent given $\mathbf{X}_L$, denoted $I(\mathbf{X}_J, \mathbf{X}_L, \mathbf{X}_K)_P$ iff $P(x_J, x_K \mid x_L) = P(x_J \mid x_L) \cdot P(x_K \mid x_L)$ for all possible values of $\mathbf{X}_J$, $\mathbf{X}_K$ and $\mathbf{X}_L$ for which $P(x_L) > 0$. $I(\mathbf{X}_J, \mathbf{X}_L, \mathbf{X}_K)_P$ is called a *(conditional independence) statement*.

The importance of $d$-separation stems from the following theorem [Verma and Pearl 1988; Geiger and Pearl 1988].

**Theorem 2:** Let $\mathbf{P}_D = \{P \mid (D, P) \text{ is a Bayesian network}\}$. Then,

$$I(J, L, K)_D \leftrightarrow I(X_J, X_L, X_K)_P \text{ for all } P \in \mathbf{P}_D$$

The "only if" part (soundness) states that whenever $I(J, L, K)_D$ holds in $D$, it must represent an independency that holds in every underlying distribution. The "if" part (completeness) asserts that any independency that is not detected by $d$-separation cannot be shared by all distributions in $\mathbf{P}_D$ and, hence, cannot be revealed by non-numeric methods.

## 3. THE MAIN RESULTS

In this section we develop a linear time algorithm for identifying the set of nodes $K$ that are $d$-separated from $J$ by $L$. The soundness and completeness of $d$-separation guarantees that the set of variables $X_K$ corresponding to the set of nodes $K$ is the maximal set of variables that can be identified as being independent of $X_J$ give $X_L$, without resorting to numerical calculations. The proposed algorithm is a variant of the well known Breath First Search algorithm; it finds all nodes reachable from $J$ through an active trail (by $L$), hence the maximal set of nodes $K$ satisfying $I(J, L, K)_D$. This task can be viewed as an instance of a more general task of finding a path in a directed graph for which some specified pairs of links are restricted not to appear consecutively. In this context, $d$-separation is viewed as a specification for such restrictions, for example, two links $u \to v$, $v \to w$ cannot appear consecutively in an active trail unless $v \in L$ or $v$ has a descendent in $L$. The following notations are employed: $D = (V, E)$ is a directed graph (not necessarily acyclic) where $V$ is a set of nodes, $E \subseteq V \times V$ is the set of (directed) links and $F \subseteq E \times E$ is a list of pairs of adjacent links that cannot appear consecutively ($F$-connotes fail). We say that an ordered pair of links $(e_1, e_2)$ is *legal* iff $(e_1, e_2) \notin F$, and that a path is *legal* iff every pair of adjacent links on it is legal. We emphasize that by "path" we mean a directed path, not a trail.

We propose a simple algorithm for the following problem: Given a finite directed graph $D = (V, E)$, a subset $F \subseteq E \times E$ and a set of nodes $J$, find all nodes reachable from $J$ via a legal path in $D$. The algorithm and its proof are a slight modification of those found in Even [1979].

### Algorithm 1

**Input:** A directed graph $D = (V, E)$, a set of illegal pairs of links $F$ and a set of nodes $J$.

**Output:** A labeling of the nodes such that a node is labeled with $R$ (connoting "reachable") iff it is reachable from $J$ via a legal path.

(i) Add a new node $s$ to $V$ and for each $j \in J$, add the link $s \to j$ to $E$ and label them with 1. Label $s$ and all $j \in J$ with $R$. Label all other nodes and links with "undefined."

(ii) $i := 1$

(iii) Find all unlabeled links $v \to w$ adjacent to at least one link $u \to v$ labeled $i$, such that $(u \to v, v \to w)$ is a legal pair. If no such link exists, stop.

(iv) Label each link $v \to w$ found in Step (iii) with $i+1$ and the corresponding node $w$ with $R$.

(v) $i := i+1$, Goto Step (iii).

**Algorithm 2**

**Input:**  A Bayesian network $D = (V, E)$ and two disjoint sets of nodes $J$ and $L$.

**Data Structure:** A list of incoming links (in-list) for each node $v \in V$.

**Output:** A set of nodes $K$ where $K = \{\alpha \mid I(J, L, \alpha)_D\}$.

(i) Construct the following table:

$$\text{descendent}[v] = \begin{cases} \text{true} & \text{if } v \text{ is or has a descendent in } L \\ \text{false} & \text{otherwise} \end{cases}$$

(ii) Construct a directed graph $D' = (V, E')$ where
$$E' = E \cup \{(u \to v) \mid (v \to u) \in E\}$$

(iii) Using algorithm 1, find the set of all nodes $K'$ which have a legal path from $J$ in $D'$, where a pair of links $(u \to v, v \to w)$ is legal iff $u \neq w$ and either 1) $v$ is a head-to-head node on the trail $u\text{—}v\text{—}w$ in $D$ and descendent$[v]$ = true or 2) $v$ is not a head-to-head node on the trail $u\text{—}v\text{—}w$ in $D$ and $v \notin Z$.

(iv) $K = V - (K' \cup J \cup L)$

Return ($K$).

The correctness of this algorithm is established by the following argument.

**Lemma 4:** For every node $\alpha \notin J \cup L$, $\alpha$ is reachable from $J$ via a legal trail in $D'$ iff there is an active path by $L$ from $J$ to $\alpha$ in $D$.

**Proof:** For $\alpha \notin J \cup L$ and $x_0 \in J$, if $(x_0 - x_1 \ldots \alpha)$ is an active trail (by $L$) in $D$, then the directed path $(x_0 \to x_1 \to \ldots \alpha)$ is a legal path in $D'$, and vise versa. (We have eliminated the case $\alpha \in J \cup L$ for technical convenience; the trail $(x_0 - x_1 \ldots \alpha)$ is not active nor non-active because, by our definition, $J$, $L$, and $\{\alpha\}$ must be disjoint. □

**Theorem 5:** The set $K$ returned by the algorithm is exactly $\{\alpha \mid I(J, L, \alpha)_D\}$.

**Proof:** The set $K'$ constructed in Step (iii) contains all nodes reachable from $J$ via a legal path in $D'$. Thus, by lemma 4, $K'$ contains all nodes not in $J \cup L$ that are reachable from $J$ via an active trail (by $L$) in $D$. However, $I(J, L, \alpha)_D$ holds iff $\alpha \notin J \cup L$ and $\alpha$ is not reachable from $J$ (by an active path by $L$), therefore, $K = V - (K' \cup J \cup L)$ is exactly the set $\{\alpha \mid I(J, L, \alpha)_D\}$. □

Next, we show that the complexity of the algorithm is $O(|E|)$ we analyze the algorithm step by step. The first step is implemented as follows: Initially mark all nodes of $Z$ with true. Follow the incoming links of the nodes in $Z$ to their parents and then to their parents and so on. This way, each link is examined at most once, hence the entire step requires $O(|E|)$ operations. The second step requires the construction of a list for each node that specifies all the links that emanate from $v$ in $D$ (out-list). The in-list and the out-list completely and explicitly specify the topology of $D'$. This step also requires $O(|E|)$ steps. Using the two lists the task of finding a legal pair in step (iii) of algorithm 2 requires only constant time; if $e_i = u \to v$ is labeled $i$ then depending upon the direction of $u - v$ in $D$ and whether $v$ is or has a descendent in $Z$, either all links of the out-list of $v$, or all links of the in-list of $v$, or both are selected. Thus, one operation per each encounted link is performed. Hence, Step (iii) requires no more than $O(|E|)$ operation which is therefore the upper bound (assuming $|E| \geq |V|$) for the entire algorithm.

The above algorithm can also be employed to verify whether a specific statement $I(J, L, K)_D$ holds in a dag $D$. Simply find the set $K_{max}$ of all nodes that are d-separated from $J$ given $L$ and observe that $I(J, L, K)_D$ holds in $D$ iff $K \subseteq K_{max}$. In fact, for this task, algorithm 2 can slightly be improved by forcing termination once the condition $K \subseteq K_{max}$ has been detected. Lauritzen at al [1988] have recently proposed another algorithm for the same task. Their algorithm consists of the following steps. First, form a dag $D'$ by removing from $D$ all nodes which are not ancestors of any node in $J \cup K \cup L$ (and removing their incident links). Second, form an undirected graph $G$, called the *moral graph*, by stripping the directionality of the links of $D'$ and

### Algorithm 2

**Input:** A Bayesian network $D = (V, E)$ and two disjoint sets of nodes $J$ and $L$.

**Data Structure:** A list of incoming links (in-list) for each node $v \in V$.

**Output:** A set of nodes $K$ where $K = \{\alpha \mid I(J, L, \alpha)_D\}$.

(i) Construct the following table:

$$\text{descendent}[v] = \begin{cases} \text{true} & \text{if } v \text{ is or has a descendent in } L \\ \text{false} & \text{otherwise} \end{cases}$$

(ii) Construct a directed graph $D' = (V, E')$ where
$$E' = E \cup \{(u \to v) \mid (v \to u) \in E\}$$

(iii) Using algorithm 1, find the set of all nodes $K'$ which have a legal path from $J$ in $D'$, where a pair of links $(u \to v, v \to w)$ is legal iff $u \neq w$ and either 1) $v$ is a head-to-head node on the trail $u\text{---}v\text{---}w$ in $D$ and descendent$[v]$ = true or 2) $v$ is not a head-to-head node on the trail $u\text{---}v\text{---}w$ in $D$ and $v \notin Z$.

(iv) $K = V - (K' \cup J \cup L)$

Return $(K)$.

The correctness of this algorithm is established by the following argument.

**Lemma 4:** For every node $\alpha \notin J \cup L$, $\alpha$ is reachable from $J$ via a legal trail in $D'$ iff there is an active path by $L$ from $J$ to $\alpha$ in $D$.

**Proof:** For $\alpha \notin J \cup L$ and $x_0 \in J$, if $(x_0 - x_1 ... \alpha)$ is an active trail (by $L$) in $D$, then the directed path $(x_0 \to x_1 \to ... \alpha)$ is a legal path in $D'$, and vise versa. (We have eliminated the case $\alpha \in J \cup L$ for technical convenience; the trail $(x_0 - x_1 ... \alpha)$ is not active nor non-active because, by our definition, $J$, $L$, and $\{\alpha\}$ must be disjoint.) $\square$

**Theorem 5:** The set $K$ returned by the algorithm is exactly $\{\alpha \mid I(J, L, \alpha)_D\}$.

**Proof:** The set $K'$ constructed in Step (iii) contains all nodes reachable from $J$ via a legal path in $D'$. Thus, by lemma 4, $K'$ contains all nodes not in $J \cup L$ that are reachable from $J$ via an active trail (by $L$) in $D$. However, $I(J, L, \alpha)_D$, holds iff $\alpha \notin J \cup L$ and $\alpha$ is not reachable from $J$ (by an active path by $L$), therefore, $K = V - (K' \cup J \cup L)$ is exactly the set $\{\alpha \mid I(J, L, \alpha)_D\}$. $\square$

Next, we show that the complexity of the algorithm is $O(|E|)$ we analyze the algorithm step by step. The first step is implemented as follows: Initially mark all nodes of $Z$ with true. Follow the incoming links of the nodes in $Z$ to their parents and then to their parents and so on. This way, each link is examined at most once, hence the entire step requires $O(|E|)$ operations. The second step requires the construction of a list for each node that specifies all the links that emanate from $v$ in $D$ (out-list). The in-list and the out-list completely and explicitly specify the topology of $D'$. This step also requires $O(|E|)$ steps. Using the two lists the task of finding a legal pair in step (iii) of algorithm 2 requires only constant time; if $e_i = u \to v$ is labeled $i$ then depending upon the direction of $u - v$ in $D$ and whether $v$ is or has a descendent in $Z$, either all links of the out-list of $v$, or all links of the in-list of $v$, or both are selected. Thus, a constant number of operations per encountered link is performed. Hence, Step (iii) requires no more than $O(|E|)$ operation which is therefore the upper bound (assuming $|E| \geq |V|$) for the entire algorithm.

The above algorithm can also be employed to verify whether a specific statement $I(J, L, K)_D$ holds in a dag $D$. Simply find the set $K_{\max}$ of all nodes that are d-separated from $J$ given $L$ and observe that $I(J, L, K)_D$ holds in $D$ iff $K \subseteq K_{\max}$. In fact, for this task, algorithm 2 can slightly be improved by forcing termination once the condition $K \subseteq K_{\max}$ has been detected. Lauritzen at al [1988] have recently proposed another algorithm for the same task. Their algorithm consists of the following steps. First, form a dag $D'$ by removing from $D$ all nodes which are not ancestors of any node in $J \cup K \cup L$ (and removing their incident links). Second, form an undirected graph $G$, called the *moral graph*, by stripping the directionality of the links of $D'$ and connecting any two nodes that have a common child (in $D'$) which is or has a descendent in $L$. Third, they show that $I(J, L, K)_D$ holds iff all (undirected) paths between $J$ and $K$ in $G$ are intercepted by $L$.

The complexity of the moral graph algorithm is $O(|V|^2)$ because the moral graph $G$ may contain up to $|V|^2$ links. Hence, checking separation in $G$ could require $O(|V|^2)$ steps. Thus, our algorithm is a moderate improvement as it only requires $O(|E|)$ steps. The gain is significant mainly in sparse graphs where $|E| = O(|V|)$. We note that if the maximal number of parents of each node is bounded by a constant, then the two algorithms achieve the same asymptotic behavior i.e., linear in $|E|$. On the other hand, when the task is to find *all* nodes $d$-separated from $J$ by $L$ (not merely validating a given independence), then a brute force application of the moral graph algorithm requires $O(|V|^3)$ steps, because for each node not in $J \cup L$ the algorithm must construct a new moral graph. Hence, for this task, our algorithm offers a considerable improvement.

The inference engine of Bayesian networks has also been used for decision analysis; an analyst consults an expert to elicit information about a decision problem, formulates the appropriate network and then by an automated sequence of graphical and probabilistic manipulations an optimal decision is obtained [Howard and Matheson 1981; Olmsted 1984; Shachter 1988]. When such a network is constructed it is important to determine the information needed to answer a given query $P(x_J | x_L)$ (where $\{J\} \cup L$ is an arbitrary set of nodes in the network), because some nodes might contain no relevant information to the decision problem and eliciting their numerical parameters is a waste of effort [Shachter 1988]. Assuming that each node $X_i$ stores the conditional distribution $P(x_i | x_{\pi(i)})$, the task is to identify the set $M$ of nodes that must be consulted in the process of computing $P(x_J | x_L)$ or, alternatively, the set of nodes that can be assigned arbitrary conditional distributions without affecting the quantity $P(x_J | x_L)$. The required set can be identified by the $d$-separation criterion. We represent the parameters of the distribution $P(x_i | x_{\pi(i)})$ as a dummy parent $p_i$ of node $i$. This is clearly a legitimate representation complying with the format of Eq. (1), since for every node $X_i$, $P(x_i | x_{\pi(i)})$ can also be written as $P(x_i | x_{pi(i)}, p_i)$, so $p_i$ can be regarded as a parent of $X_i$. From Theorem 1, all dummy nodes that are $d$-separated from $J$ by $L$ represent variables that are conditionally independent of $J$ given $L$ and so, the information stored in these nodes can be ignored. Thus, the information required to compute $P(x_J | x_K)$ resides in the set of dummy nodes which are not $d$-separated from $J$ given $L$. Moreover, the completeness of $d$-separation further implies that $M$ is minimal; no node in $M$ can be exempted from processing on purely topological grounds (i.e., without considering the numerical values of the probabilities involved). The algorithm below summarizes these considerations:

### Algorithm 3

**Input:** A Bayesian network, two sets of nodes $J$ and $L$.

**Output:** A set of nodes $M$ that contains sufficient information to compute $P(x_j | x_L)$

(i) Construct a dag $D'$ by augmenting $D$ with a dummy node $v'$ for every node $v$ in $D$ and adding a link $v' \to v$.

(ii) Use algorithm 2 to compute the set $K'$ of nodes not $d$-separated from $J$ by $L$.

(iii) Let $M$ be the set of all dummy nodes $v'$ that are included in $K'$.

We conclude with an example. Consider the network $D$ of Figure 3 and a query $P(x_3)$.

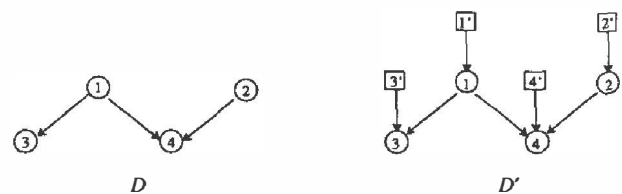

Figure 3

The computation of $P(x_3)$ requires only to multiply the matrices $P(x_3|x_1)$ and $P(x_1)$ and to sum over the values of $X_1$. These two matrices are stored at the dummy nodes 1' and 3', which are the only dummy nodes not $d$-separated from node 3 (given $\varnothing$). Thus, algorithm 3 reveals the fact that the parameters represented by node 2' and 4' ($P(x_2), P(x_4 | x_1, x_2)$) are not needed for the computation of $P(x_3)$. Note, however, that knowing the value of $X_4$ might influence the computation of $P(x_3)$, because $X_3$ and $X_4$ could be dependent. The value of $X_2$, on the other hand, never affects this computation because $X_2$ is independent of $X_3$. This example shows that the question of whether a variable influences the computation of a query and the question of whether the parameters stored with that variable influence the same query are two different questions. Algorithm 3, by representing parameters as dummy variables, solves the latter problem by transforming it to an instance of the former.

Shachter was the first to address the problem of identifying irrelevant parameters [Shachter 1988][3]. Our formulation provides several advantages. First, we distinguish between sensitivity to variable instantiations and sensitivity to parameter values, and the algorithm we provide can be tailored to solve either one of these problems. Shachter's algorithm handles the second problem and, therefore, it does not reveal all the independencies that are implied by the topology of the dag. For example, in Figure 3, Shachter's algorithm would correctly conclude that nodes 2 and 4 both contain no relevant information for the computation of $P(x_3)$. Yet, $X_2$ is independent of $X_3$, while $X_4$ and $X_3$ might be dependent, a distinction not addressed in Shachter's algorithm. Second, our method is comprised of two components, 1) declarative characterization of the independencies encoded in the network (i.e., the $d$-separation the criterion) and 2) procedural implementation of the criterion defined in 1). This approach facilitates a clear proof of the validity and maximality of the graphical criterion, independent of the details of the algorithm, followed by proofs of the algorithm's correctness and optimality ( it requires only $O(|E|)$ steps). In Shachter's treatment the characterization of the needed parameters is inseparable from the algorithm, hence, it is harder to establish proofs of correctness and maximality.

## ACKNOWLEDGEMENT

We thank Eli Gafni for his help in developing algorithm 1, and to Azaria Paz and Ross Shachter for many stimulating discussions.## REFERENCES

S. Even. 1979. *Graph Algorithms*, Computer Science Press.

E. Gafni. Personal communication, 1988.

D. Geiger & J. Pearl. August 1988. "On The Logic of Causal Models," *Proc. of the 4th Workshop on Uncertainty in AI*, St. Paul, Minnesota, pp. 136-147.

D. Geiger, T. S. Verma and J. Pearl. 1989. "Identifying independence in Bayesian networks", Technical report R-116-I, UCLA Cognitive Systems Laboratory. In preparation.

C. Glymour, R. Scheines, P. Spirtes and K. Kelly. 1987. *Discovering Causal Structure*, New York: Academic Press.

R. A. Howard & J. E. Matheson. 1981. "Influence Diagrams," In, *Principles and Applications of Decision Analysis*, Menlo Park, CA: Strategic Decisions Group.

S.L. Lauritzen, D. Spiegelhalter. 1989. "Local computations with probabilities on graphical structures and their applications to expert systems." *J. Royal Statist. Soc.*, ser. B.

S.L. Lauritzen, A.P. Dawid, B.N. Larsen and H.G. Leimer. October 1988. "Independence Properties of Directed Markov Fields," Technical Report R 88-32, Aalborg Universitetscenter, Aalborg Denmark.

S. M. Olmsted. 1983. "On Representing and Solving Decision Problems," Ph.D. Thesis, EES Dept., Stanford University.

J. Pearl. 1988. *Probabilistic Reasoning in Intelligent Systems: Networks of Plausible Inference*, San Mateo, CA: Morgan Kaufmann.

R.D. Shachter. 1988. "Probabilistic Inference and Influence Diagrams," *Operations Research*, Vol. 36, pp. 589-604.

J.Q. Smith. June 1987. "Influence Diagrams for Statistical Modeling," Technical report #117, department of Statistics, University of Warwick, Coventry, England.

T. Verma & J. Pearl. August 1988. "Causal Networks: Semantics and Expressiveness," *Proceedings of the 4th*---

(3) Shachter also considers deterministic variables which we treat in [Geiger at al. 1989].

124